\documentclass[runningheads]{llncs}
\usepackage{graphicx}
\begin{document}
\title{Cyrus 2D Simulation Team Description Paper 2018}
%
%
\author{Nader Zare\inst{1} \and
Mohsen Sadeghipour\inst{2} \and
Ashkan Keshavarzi\inst{3} \and
Mahtab Sarvmaili\inst{1} \and
Amin Nikanjam\inst{1} \and
Reza Aghayari \and
Arad Firouzkoohi \and
Mohammad Abolnejad \and
Sina Elahimanesh \and
Amin Akhgari
}
\authorrunning{N. Zare et al.}
%
\institute{K. N. Toosi University of Technology\\
\email{nader.zare88@gmail.com, mahtab.sarvmaili@gmail.com, nikanjam@kn2c.ac.ir}\\
\and
 Qazvin Azad University\\
\email{mohsen.sadeghipour@icloud.com} \\
\and Tehran University \\
\email{keshavarzi.a@ut.ac.ir}
}
\maketitle              
\begin{abstract}
Cyrus 2D Soccer Simulation was established 2012 with the aim of research and develop in multi agents systems. This year we have joined with
Ziziphus for collaboration and speed up our researches. This paper express a brief
description of a method for predicting player's behavior in a multi agent system
using neural network with a dataset in three level (low, mid, high). The dataset
was obtained from log files of past years RoboCup's matches. Behavior Prediction is used in block, mark and defensive decisions. The base code that Cyrus
used is agent 3.11\cite{ref1}.

\end{abstract}
\section{Introduction}
Cyrus\cite{ref2}robotic team was established in 2012 with the aim of research and development in fields of artificial intelligence, multi-agent decision, and deep learning. The founder of this team was the students of Shiraz University of technology but nowadays Cyrus keeps on its activities under Atomic Energy High school. Current team members are formed from the students of the K. N. Toosi University of Technology, Tehran University, Qazvin Azad University and Atomic Energy High School. Dr. Amin Nikanjam, assistant professor in the K. N. Toosi University of Technology, is the advisor of the team. In 2018, Cyrus joined with Ziziphus\cite{ref3}. Since 2013, Cyrus has taken part in Global RoboCup's Competition and in these years won 5th, 8th, 9th, 12th, and 4th places. We also took part in Iran Open Competitions since 2013 and Cyrus won first place in 2014 and challenge's first place in 2017. Cyrus won first place in Shiraz Open two times. Currently, our main focus is on deep learning algorithms, reinforcement learning algorithms and their application in 2D Soccer simulation. In this paper, we will talk about predicting kickable player's behavior.

\section{Players Behavior Prediction}
The main purpose of the soccer games and 2D Soccer Simulation league is to propel the ball into the opponents' goal. For this purpose, the players who get the ball should make a way to the penalty area by using pass, dribble, and finally make a shoot to the goal. Predicting this behavior have many advantages will be discussed in next section. To predict the player’s behavior, we use the neural network and to train this network we generated a dataset from the RoboCup's teams and fed the neural network by this datasets. In the dataset the state of the game is categorized in three level, low-level, mid-level and high level, the label for each data is equal to the receiver of the ball. After training a neural network with high-level dataset for Helios2017 team, we can predict the target player with 90 percent probability out of two players with the high score.
\subsection{A Subsection Sample}
Predicting players' behavior is applicable in\\
1. Improving Team's defensive tactics by predicting opponent's behavior: The opponent behavior prediction is helpful in defense area because it could help to organize the players in defending zone.\\
2. Improving Team's offensive tactics by predicting opponent's behavior: Our team can mimic the decisioning of powerful teams in attack time, by using the prediction of their offensive behavior.\\
3. Improving Team's unmarking process by predicting teammate's behavior: our players can predict his teammate's behavior during offense situation and use this prediction to track the chain action of kickable player and the ball’s path. This prediction provide successful result while players making unmark decision.\\
Using opponent's behavior in offense strategy already implement in Cyrus team and all the improvement will be implemented until RoboCup.
\subsection{Generating offline dataset}
To train the estimator neural network in three levels we need to build datasets in three levels. To generate such a dataset, we used an open source Python log extractor\cite{ref4}. We generate a dataset so that every data is composed of two parts, first, the input part of the data defines the match state, second part is an 11-dimensional output, and each one of these dimensions is assigned to a player and can be used to represent the target player in the current state. The following subsections introduce the low-level state and action space used by agents in this domain.
\subsubsection{Low-Level, Mid-Level, High-Level State Space}
The input part in low level, mid-level and in high-level state spaces contains 92,352 and 385 features respectively. The input part in all levels is shown in Table~\ref{tab1}

\begin{table}
\centering
\caption{Feature and Dimension for input part of dataset in each level.}\label{tab1}
\begin{tabular}{|l|l|l|l|l|}
\hline
Feature & \rotatebox{45}{Numbers of Dimension} & \rotatebox{45}{Low-Level} & \rotatebox{45}{Mid-Level} & \rotatebox{45}{High-Level}\\
\hline
Position of Ball & 2 & Yes & Yes & Yes \\
Velocity of Ball & 2 & Yes & Yes & Yes \\
Position of Players & 44 & Yes & Yes & Yes \\
Velocity of Player  & 44 &Yes  & Yes & Yes \\
Distance of the Ball To Important Points of the Field & 9 & No & Yes & Yes \\
Angle of the Ball To Important Point of Field & 9 & No & Yes & Yes \\
Distance of the Ball To Players & 22 & No & Yes & Yes \\
Angle of the Ball To Players & 22 & No & Yes & Yes \\
Free Angle of the Ball To Teammate & 11 & No & No & Yes \\
Distance of Players To Important Point of Field & 198 & No & Yes & Yes \\
Minimum Distance of Players to Teammate & 11 & No & No & Yes \\
Minimum Distance of Players to Opponent & 11 & No & No & Yes \\
\hline
{\bfseries Number of Dimension} & {\bfseries 385} & {\bfseries 92} & {\bfseries 352} & {\bfseries 385} \\
\hline
\end{tabular}
\end{table}

\begin{figure}
\includegraphics[width=\textwidth]{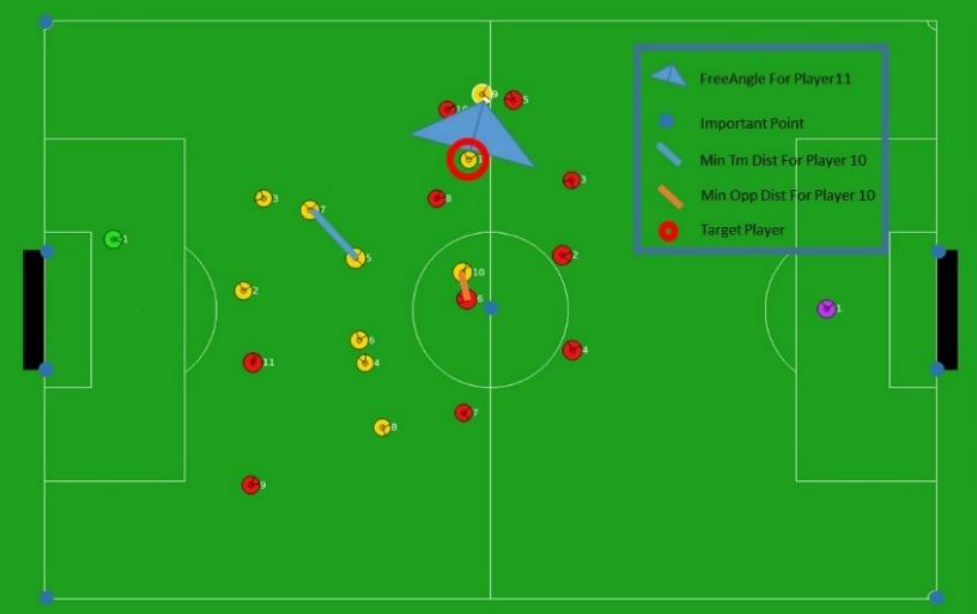}
\caption{Overview of the training data features (input part).} \label{fig1}
\end{figure}

\subsection{Training Neural Network}
We have implemented rough neural network using "Object Oriented Neural Network"\cite{ref5} in Cyrus2017 source code. To train the estimator neural network we also using this rough neural network. Objective oriented Neural Network is implemented in C++ and it's under open source license. 
To generate the best sample of training data, 200 matches were played between Helios\cite{ref6}, Gliders2016\cite{ref7} and other teams from World Championships. We extracted 50000 data for training Neural Networks in three modes of low-level, mid-level and high-level in 200 games between Helios2017 and other teams. The following diagram is related to the training of the neural network in three modes that has already been discussed in section 2-2.

\begin{figure}
\includegraphics[width=\textwidth]{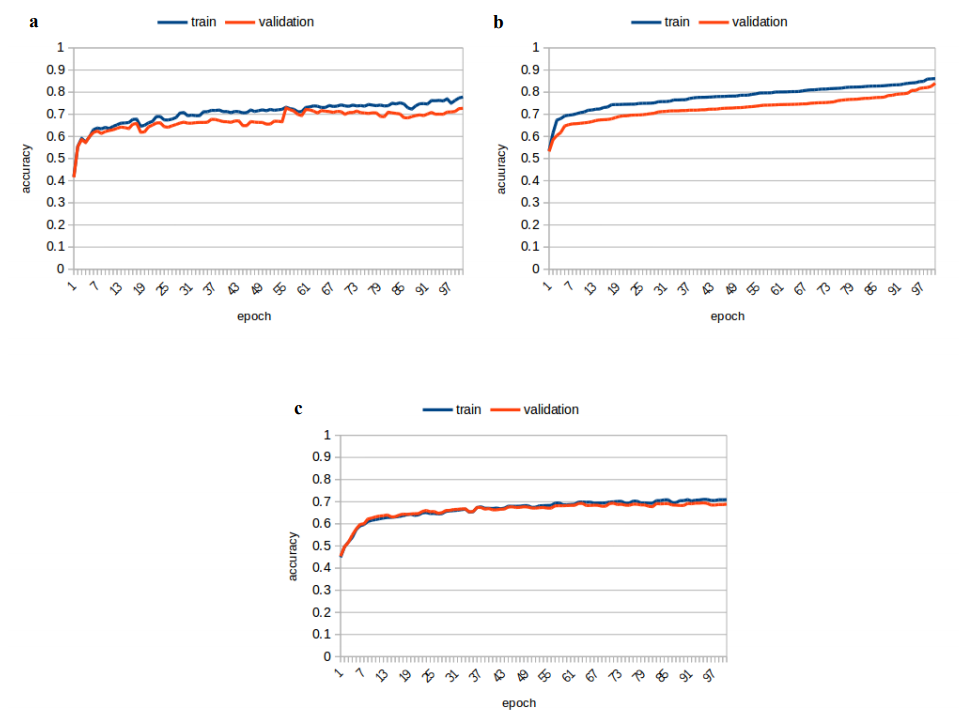}
\caption{Accuracy of neural network in predicting target player.} \label{fig2}
\end{figure}

The best result of the neural network given those three types of the training data for Glider2016, Helios2017 and Cyrus is shown in the following Table~\ref{tab2}. According to these results, we decide to use high-level dataset.

\begin{table}
\centering
\caption{Shows the accuracy of Behavior Prediction for Gliders, Helios, and Cyrus in each dataset's level.}\label{tab2}
\begin{tabular}{|l|l|l|l|}
\hline
Team’s Name & Low-Level & Mid-Level & High-Level\\
\hline
Gliders & 78\%  & 82\% & 89\% \\
Helios & 81\%  & 84\% & 90\% \\
Cyrus &  74\% & 81\% & 85\% \\
\hline
\end{tabular}
\end{table}

\subsection{Using this dataset in Cyrus team and Final Results}
We use “opponent's behavior prediction” in the defensive strategy of the Cyrus. In this section, we are going to details of the algorithm and the results of the implemented neural network. In defensive strategy of our team, each one of our players will score the opponents. The scoring formula is shown in the Equation.1 and Equation.2.

\begin{equation} \label{eqn1}
{OpponentScore(state,unum)=OppPosition.X + max(0.40 - Dist(OppPosition.CyrusCenterGoal)}
\end{equation}
\begin{equation} \label{eqn2}
{OpponentScore(state,unum)=OpponentScore(state,unum)*(1+NeuralNetworkOutput(unum))} 
\end{equation}

We calculate the score of opponents based on:\\
1. The closeness of the player to the goal \\
2. The probability of passing the ball to this player from kickable player.\\
After calculating this formula for each opponent, and sort them according to their score. Then we will assign mark and block behavior score to each "opponent-teammate" pair in a 2dimensional structure, then we use greedy algorithm to select our best player to mark or to block the opponent who has highest Opponent-Score, then we remove best player and marked or blocked opponent from 2dimensional structure and then we re-run the explained algorithm. We have used the above algorithm without using “opponent behavior prediction” in Cyrus2017, in Table~\ref{tab3} showed the result of Cyrus2017 and Cyrus2018 with other teams.

\begin{table}
\centering
\caption{Shows the average results of Cyrus2017 and Cyrus2018 against Helios2017,
Glider2016, and Oxsy2017.}\label{tab3}
\begin{tabular}{|l|l|l|}
\hline
Teams &Cyrus2017 &Cyrus2018\\
\hline
Helios2017 & 3.1 – 0.8 & 2.1 – 0.9 \\
Glides2016 & 3.7 – 1.1 & 1.89 – 1.3 \\
Oxsy2017 & 2.8 – 0.75 & 1.39 – 1.1 \\
\hline
\end{tabular}
\end{table}

\subsection{Future work}
In our future works, we are planning to use deep reinforcement learning alongside of behavior prediction. We consider training the behavior of kickable opponent for each players and improve their deaccessioning process by using actor-critic [8] algorithm. To improve the results of the behavior prediction we want to partition the field into some geometric shapes and feed these shapes to a convolutional network to predict the behavior of players.

\end{document}